%% file: sample-base.tex
\documentclass[sigconf]{acmart}

\AtBeginDocument{%
  \providecommand\BibTeX{{%
    \normalfont B\kern-0.5em{\scshape i\kern-0.25em b}\kern-0.8em\TeX}}}

\usepackage{algorithm}
\usepackage{algorithmic}

\usepackage{subcaption}
\usepackage{multirow}
\usepackage{indentfirst}

\usepackage{makecell}
\usepackage{bbding}
\usepackage{diagbox}
\usepackage{color}

\usepackage{balance}

\usepackage{wrapfig}
\newcommand{\red}[1]{\textcolor{red}{#1}}
\newcommand{\blue}[1]{\textcolor{blue}{#1}}
\newcommand{\revised}[1]{{\color{black} #1}}

\copyrightyear{2023}
\acmYear{2023}
\setcopyright{acmlicensed}\acmConference[MM '23]{Proceedings of the 31st ACM International Conference on Multimedia}{October 29-November 3, 2023}{Ottawa, ON, Canada}
\acmBooktitle{Proceedings of the 31st ACM International Conference on Multimedia (MM '23), October 29-November 3, 2023, Ottawa, ON, Canada}
\acmPrice{15.00}
\acmDOI{10.1145/3581783.3611876}
\acmISBN{979-8-4007-0108-5/23/10}

\begin{document}

\title{EasyNet: An Easy Network for 3D Industrial Anomaly Detection}
\author{Ruitao Chen}
\authornote{Equally contribute to this work}
 \affiliation{%
  \institution{Southern University of Science and Technology}
  \country{Shenzhen, China}}
 \email{chenrt2022@mail.sustech.edu.cn}
 
 \author{Guoyang Xie}
 \authornotemark[1]
 \affiliation{%
  \institution{Southern University of Science and Technology}
  \country{Shenzhen, China}}
      \affiliation{%
  \institution{University of Surrey}
  \country{Guildford GU2 7XH, UK}
  }
 \email{guoyang.xie@surrey.ac.uk}

 \author{Jiaqi Liu}
\authornotemark[1]
 \affiliation{%
  \institution{Southern University of Science and Technology}
  \country{Shenzhen, China}}
 \email{liujq32021@mail.sustech.edu.cn}

 \author{Jinbao Wang}
\authornote{Corresponding author}
 \affiliation{%
  \institution{Southern University of Science and Technology}
  \country{Shenzhen, China}}
 \email{linkingring@163.com}

  \author{Ziqi Luo}
 \affiliation{%
  \institution{Southern University of Science and Technology}
  \country{Shenzhen, China}}
 \email{luozq2022@mail.sustech.edu.cn}

 \author{Jinfan Wang}
 \affiliation{%
  \institution{Southern University of Science and Technology}
  \country{Shenzhen, China}}
  \affiliation{%
  \institution{Linkinsense}
  \country{Shenzhen, China}
  }
 \email{wangjf@sustech.edu.cn}
 
 \author{Feng Zheng}
 \authornotemark[2]
 \affiliation{%
  \institution{CSE and RITAS, Southern University of Science and Technology}
  \country{Shenzhen, China}}
 \email{f.zheng@ieee.org}


\renewcommand{\shortauthors}{Ruitao Chen et al.}

\begin{abstract}

3D anomaly detection is an emerging and vital computer vision task in industrial manufacturing (IM). Recently many advanced algorithms have been published, but most of them cannot meet the needs of IM. There are several disadvantages: i) difficult to deploy on production lines since their algorithms heavily rely on large pretrained models; ii) hugely increase storage overhead due to overuse of memory banks; iii) the inference speed cannot be achieved in real-time. To overcome these issues, we propose an easy and deployment-friendly network (called EasyNet) without using pretrained models and memory banks: firstly, we design a multi-scale multi-modality feature encoder-decoder to accurately reconstruct the segmentation maps of anomalous regions and encourage the interaction between RGB images and depth images; secondly, we adopt a multi-modality anomaly segmentation network to achieve a precise anomaly map; thirdly, we propose an attention-based information entropy fusion module for feature fusion during inference, making it suitable for real-time deployment. Extensive experiments show that EasyNet achieves an anomaly detection AUROC of 92.6\% without using pretrained models and memory banks. In addition, EasyNet is faster than existing methods, with a high frame rate of 94.55 FPS on a Tesla V100 GPU.
\end{abstract}


\maketitle

\section{Introduction}\label{sec:introduction}

\begin{figure}[th]
    \centering
    \includegraphics[width=0.9\linewidth]{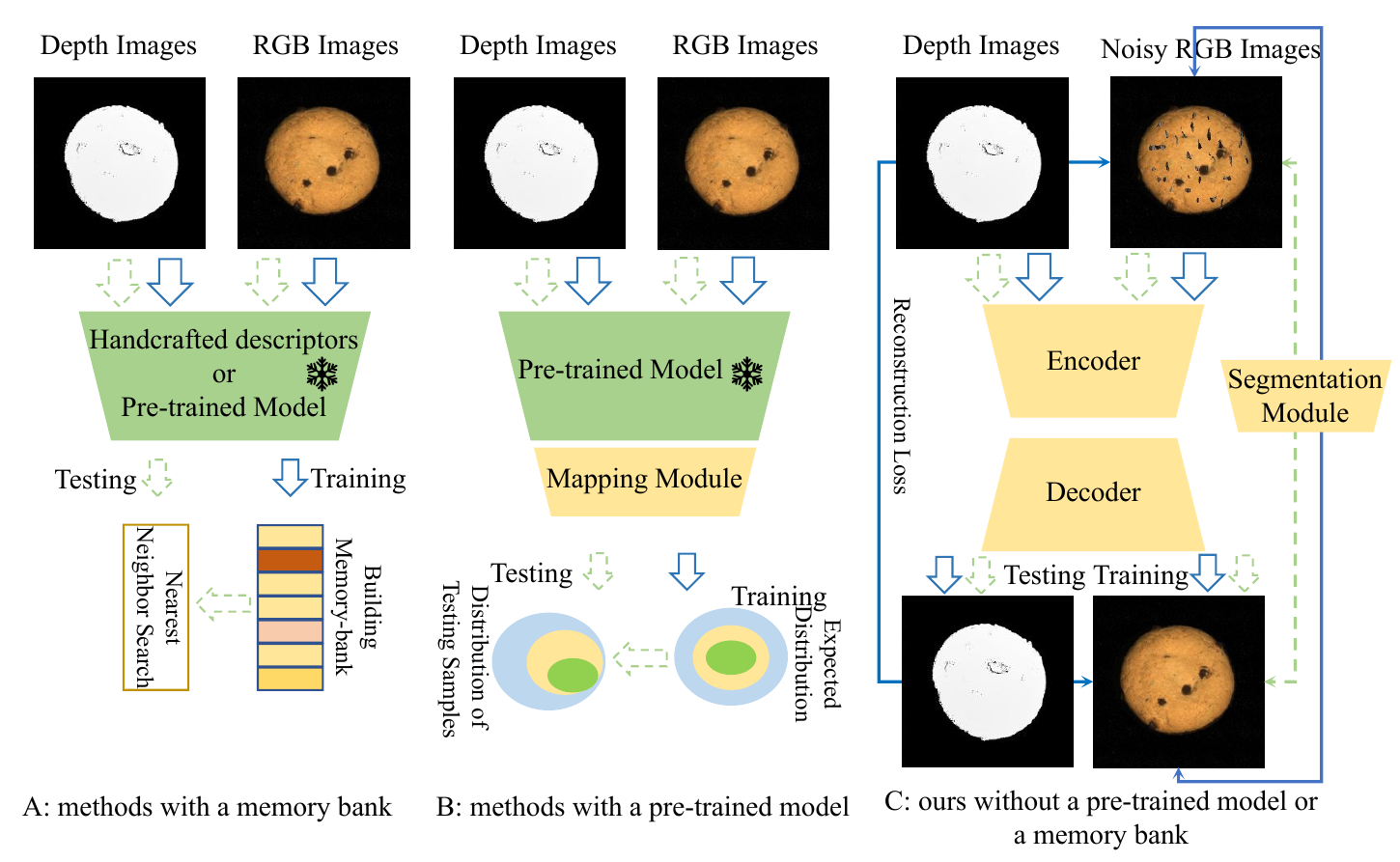}
	\caption{Illustration of 3D anomaly detection paradigms, including (A) feature embedding-based memory bank, (B) pretrained network, and (C) encoder-decoder (ours).
	}\label{fig:easynet-motivation}
\end{figure}

There is a strong need to propose a deployment-friendly 3D unsupervised anomaly detection (3D-AD) model to tap the gap, which brings 3D-AD's capabilities into the factory floor. Currently, most of anomaly detection methods~\cite{li2022towards, xie2023pushing, Xie2023IMIADII, liu2023deep} are based on 2D images. But in the quality inspection of industrial products, human inspectors utilize both color (RGB) characteristics and depth information to determine whether it is a defective product, where depth information is essential for anomaly detection. As shown in Figure~\ref{fig:3DAD-motivation}, for foam and peach, it is difficult to identify the anomalies from the RGB image alone. Though 3D-AD algorithms~\cite{Wang2023MultimodalIA, Rudolph2022AsymmetricSN, Bergmann2022AnomalyDI} are attracting interest from the academy, most of them are far from satisfactory for industrial manufacturing (IM). According to Figure~\ref{fig:easynet-motivation}, there are several issues: i) The cutting-edge 3D-AD methods steadily rely on the representational abilities of large pretrained models, leading to slow inference speed and huge storage overhead. ii) Feature embedding-based 3D-AD methods excessively use memory banks, leading to huge memory bank costs in real-world applications. Because of this, it is important and urgent to build an application-oriented 3D-AD model to meet the demands of IM. 

To avoid using large pretrained models and memory banks, we propose an easy but effective multi-modality anomaly detection and localization network, called \textbf{EasyNet}. Specifically, EasyNet consists of two parts, the Multi-modality Reconstruction Network (MRN) and the Multi-modality Segmentation Network (MSN). First, instead of using pretrained features directly, we generate synthesized anomalies on RGB images and depth images, reconstruct the original images with semantically reasonable free content, and obtain multi-scale features. At the same time, in order to simplify the anomaly detection process based on memory bank, we input the abnormal and reconstructed multi-scale features into a simple MSN to obtain an anomaly map. As shown in Figure \ref{fig:easynet-total-arch}, the entire architecture, including MRN and MSN, significantly encourages interaction between RGB and depth features.

To reduce the disturbance between RGB and depth images, we propose an attention-based information entropy fusion module. We find that some 3D-AD methods, like AST~\cite{rudolph2022asymmetric} and BTF~\cite{horwitz2022back} cannot fully utilize the advantage of multi-modality fusion, i.e., RGB-D performance is not competitive than RGB performance. The main reason is that there are no uniform abnormal patterns in RGB or depth images. For example, some anomalies can be detected by pure RGB images and depth information works as the noise and may degrade the overall anomaly detection performance. Hence, we propose a dynamic multi-modality fusion scheme to make use of RGB and depth features. The architecture of the fusion scheme is shown in Figure~\ref{fig:information_entropy_scheme}. Moreover, as shown in Table~\ref{tab:inference-accuracy}, our proposed fusion scheme is simple and much more computationally efficient than the aforementioned 3D-AD models~\cite{horwitz2022back, rudolph2022asymmetric, Wang2023MultimodalIA}. Our proposed attention-based information entropy fusion module is easy to train and apply, with outstanding performance and inference speed. As a result, EasyNet can achieve 92.6\% on MVTec 3D-AD and 86.9\% on Eyescandies in I-AUROC while running at 94.55 FPS, surpassing the previous best-published 3D-AD methods on accuracy and efficiency.

Our contributions can be summarized as follows:
\begin{itemize}
    \item EasyNet is easy to implement and deploy for 3D unsupervised anomaly detection, i.e., eliminating the usage of pretrained models and memory banks, and achieves the fastest inference speed than the existing methods, with a high frame rate of 94.55 FPS on a Tesla V100 GPU.
    \item We propose an Attention-based Information Entropy Fusion Module to integrate the image features of the multi-modal characteristics well.
    \item We propose a Multi-modality Reconstruction Network(MRN) to accurately reconstruct the anomalous region and encourage the interaction of RGB and depth.
    \item We propose a Multi-modality Segmentation Network(MSN) to output the anomaly map precisely.
    \item EasyNet obtains the state-of-the-art result in Pure RGB. Note that EasyNet obtains the best anomaly detection I-AUROC of 92.6\% in RGBD.
\end{itemize}

\section{Related Work}
Anomaly detection (AD) is a classical topic, which aims to distinguish normal samples and abnormal samples. Existing experimental settings usually only take normal samples as the training set, and evaluate the ability of the model to distinguish abnormal samples in the test set. The current unsupervised AD can be mainly divided into feature extraction-based methods and image reconstruction-based methods. The former is restricted by the pretrained model, while the latter is free from this limitation. Based on this idea, we design a reconstructive AD algorithm for RGB-D data, removing the restrictions on pretrained models and memory banks.

\subsection{2D Anomaly Detection}
Since the emergence of the MVTec AD dataset~\cite{bergmann2019mvtec}, research on AD in industrial 2D images has received more attention. Most existing research is based on this set for unsupervised AD tasks. 

There is more research on feature embedding-based methods than reconstruction-based methods. The most basic idea is to regard AD as a one-class classification problem and turn the AD problem into a problem of finding boundaries for classification. CutPaste~\cite{li2021cutpaste} and SimpleNet~\cite{liu2023simplenet} are representative methods. They make abnormal samples and change unsupervised AD datasets into supervised datasets. Teacher-student architecture is another useful approach. The teacher network distills knowledge to the student network by extracting features from normal samples. While the teacher network and the student network perform differently when producing abnormal samples and they detect anomalies through this characteristic~\cite{bergmann2020uninformed, Deng2022AnomalyDV}. Normalizing flow methods map samples into a Gaussian distribution, while abnormal samples deviate from this distribution~\cite{rudolph2021same, gudovskiy2022cflow}. Methods based on memory banks are simple but effective, whose ideas come from the k-nearest neighbors (KNN) algorithm. They store features of normal samples and calculate the distance between the features of test samples and the features of normal samples during testing to determine whether the samples are abnormal~\cite{defard2021padim, roth2022towards}.
As for reconstruction-based methods, most of them are similar in structure. They synthesize abnormal samples and restore abnormal samples to normal samples.  For example, DRAEM~\cite{zavrtanik2021draem} and NSA~\cite{schluter2022natural} synthesize abnormal samples in image level, while DSR~\cite{zavrtanik2022dsr} and UniAD~\cite{you2022unified} synthesize abnormal samples in feature level.

Generally, most of the 2D-AD methods use the pretrained model of natural images to extract RGB's features while they don't process depth information, so it is difficult to apply to 3D-AD directly. There is a certain gap between these two, and our method tries to get rid of this dependence so that 2D-AD can smoothly transition to 3D-AD.

\subsection{3D Anomaly Detection}
\begin{figure}[t]
    \centering
    \includegraphics[width=0.82\linewidth]{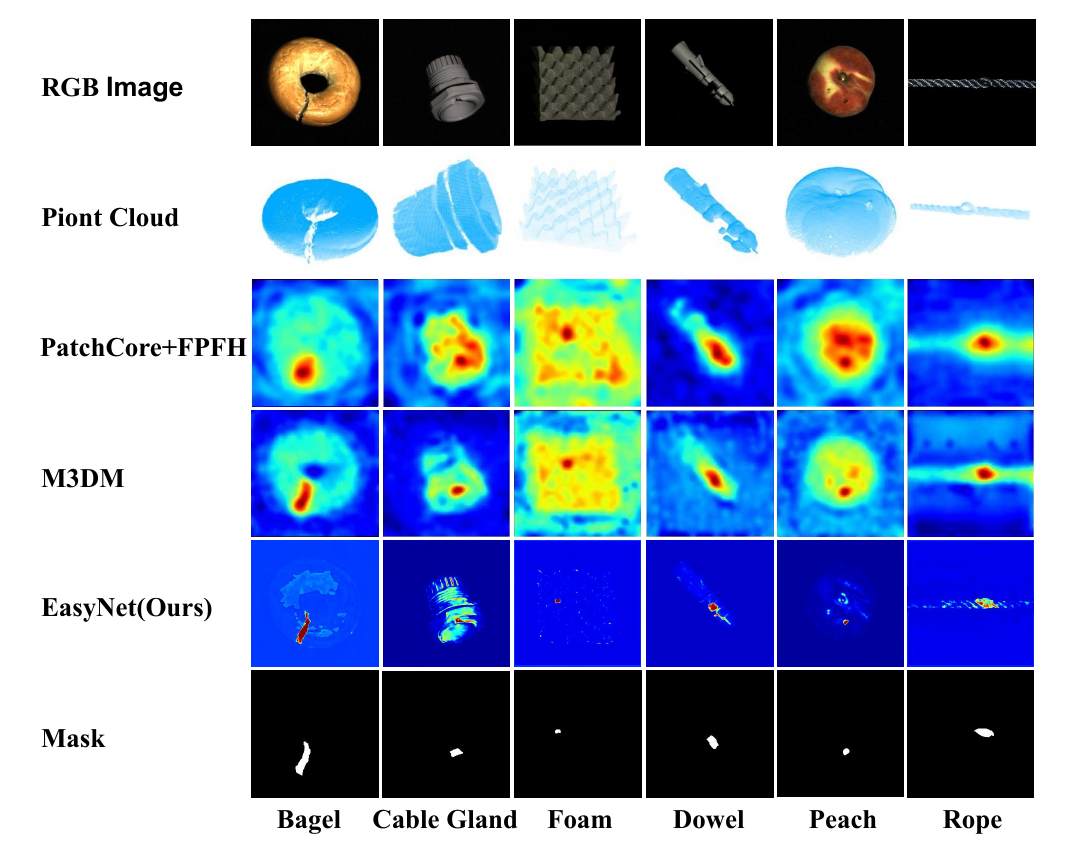}
	\caption{Simple examples of 3D anomaly detection on MVTec 3D-AD~\cite{Bergmann2021TheM3}. The first and second rows are RGB images and point clouds, and the third to fifth rows show predicted results of PatchCore+FPFH~\cite{Horwitz2022AnEI}, M3DM~\cite{Wang2023MultimodalIA}, and EasyNet (ours), respectively. Ground Truth refers to the anomaly regions.
	}\label{fig:3DAD-motivation}
\end{figure}

Different from 2D-AD, 3D-AD is a new research topic since the publication of MVTec 3D-AD~\cite{Bergmann2021TheM3}. As shown in Figure~\ref{fig:3DAD-motivation}, 3D-AD is a more challenging but also more promising research direction. The effective use of depth information can greatly improve detection accuracy in specific scenarios. On the other hand, how to integrate depth information and prevent it from interfering with RGB information is the current difficulty.

Bergmann~\textit{et al.}~\cite{Bergmann2022AnomalyDI} introduce a point-cloud feature extraction network of the teacher-student model. During training, the features extracted by the student network and the teacher network are forced to be consistent. During the test, the differences between the features extracted by the teacher-student model are compared to locate anomalies. Horwitz~\textit{et al.}~\cite{horwitz2022back} combine hand-crafted 3D descriptors with the KNN framework, a classic AD approach. These two methods are efficient, but with poor performance. AST~\cite{Rudolph2022AsymmetricSN} gets a better result in MVTec 3D-AD. However, it only uses depth information to remove the background and still uses the 2D-AD method to detect anomalies and the depth information about items is ignored. Similar to BTF, but M3DM~\cite{Wang2023MultimodalIA} extracts features from point clouds and RGB images, respectively, and fuses them to make a decision, which has a better performance than treating RGB and depth as six-channel images as BTF. The visualization effect of M3DM is shown in the fourth row of Figure~\ref{fig:3DAD-motivation}. CPMF~\cite{cao2023complementary} also adopts the KNN paradigm, but the difference lies in the fact that the authors project the point cloud from different angles into 2D images and fuse the 2D image information obtained for detection.

In summary, existing 3D-AD models either suffer from poor performance or reliance on pretrained models and memory banks. In contrast, EasyNet is simple, effective, and without relying on pretrained models or memory banks. It achieves SOTA performance outperforming all previous methods without pre-training.

\begin{figure*}[th]
    \centering
    \includegraphics[width=0.78\linewidth]{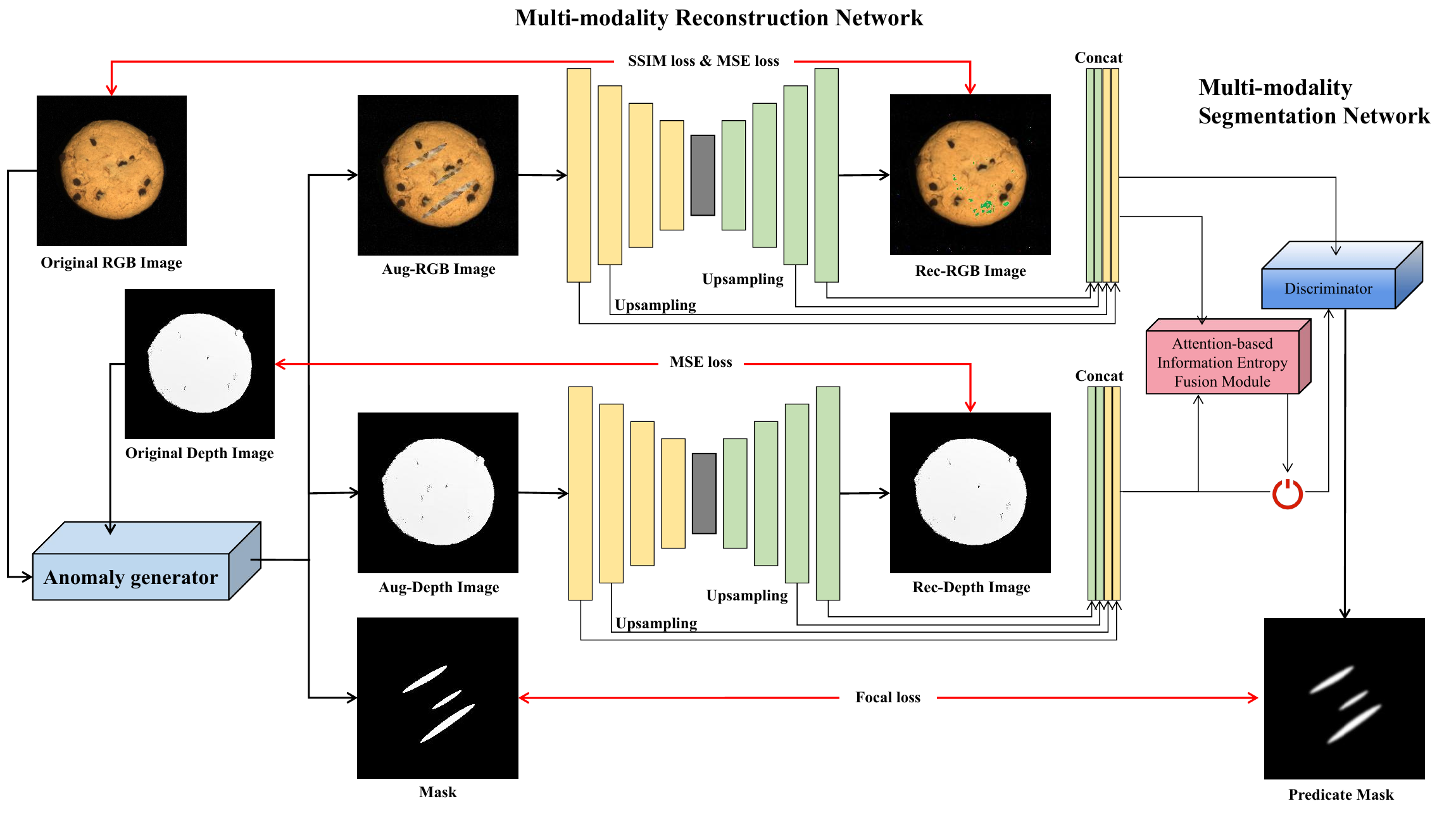}
	\caption{Total architecture of EasyNet. EasyNet consists of three main components. (1) Anomaly Generator adds Berlin noise to the original multi-modal images to simulate synthetic abnormal and normal images (RGB and depth). (2) The Multi-modality Reconstruction Network (MRN) constructs a reconstruction task to restore the enhanced synthetic anomaly images to RGB and depth images without anomalies while obtaining multi-modal feature information from multiple layers, where two layers are used as an example. (3) The Multi-modality Segmentation Network (MSN) fuses the extracted multi-modal features by utilizing an attention-based information entropy fusion module, which is fully open during training and takes control of feature flow by calculating the self-attention information entropy score for feature information from multiple modes during inference. For the reconstruction task, we use SSIM loss~\cite{Wang2004ImageQA} and MSE loss to calculate the reconstruction loss for RGB images, while only using MSE loss for depth images. Moreover, Focal loss~\cite{Lin2017FocalLF} is used to calculate the pixel classification task loss.} \label{fig:easynet-total-arch}
\end{figure*}

\section{Approach}\label{sec:method}

\subsection{Problem Definition and Challenges}
\label{subsec:challenges}

Our 3D-AD setting is similar to M3DM~\cite{Wang2023MultimodalIA} and AST~\cite{rudolph2022asymmetric} and can be formally stated as follows. Given a set of training examples $\mathcal{T} = \left\{ t_{i}\right\}_{i=1}^{N}$, in which $\left\{ t_{1}, t_{2}, \cdots, t_{N}\right\}$ are the normal samples and each of them consists of paired images, RGB image $I_{rgb}$ and depth image $I_{depth}$. In addition, $\mathcal{T}_{n}$ belongs to a certain category, $c_{j}$, $c_{j} \in \mathcal{C}$, where $\mathcal{C}$ denotes the set of all categories. During testing, given a normal or abnormal sample from a target category $c_{j}$, the AD model should predict whether or not the test 3D object is anomalous and localize the anomaly region if the anomaly is detected.

The following are the main challenges. (1) Information on normal samples is limited, each category's training dataset only contains normal samples, i.e., no pixel-level annotations of $I_{rgb}$ and $I_{depth}$. (2) It is difficult to find an effective multi-modality fusion way for anomalies that may appear in RGB, depth, or both. Simply fusing these features may negatively impact overall AD performance. (3) Real-world applications have limited storage space, so it is impractical to build a model that uses large pretrained models and memory banks.

\subsection{EasyNet}
This section provides a complete description of EasyNet. As illustrated in Figure~\ref{fig:easynet-total-arch}, the proposed model comprises a multi-scale Multi-modality Reconstruction Network (MRN), a multi-scale Multi-modality Segmentation Network (MSN) and an attention-based information entropy fusion module, with the fusion network being exclusively applied during reasoning stages. The following sections elaborate on the design and functionality of each module.

\subsubsection{Multi-modality Reconstruction Network (MRN)}
The multi-modality reconstruction network establishes a task of image reconstruction. In this task, the network reconstructs the original image from an artificially corrupted image obtained from the simulator. The network is designed as an encoder-decoder structure to transform the local features of the input image into a mode that more closely resembles the normal sample distribution.

\begin{figure}[thbp]
    \centering
    \includegraphics[width=0.9\linewidth]{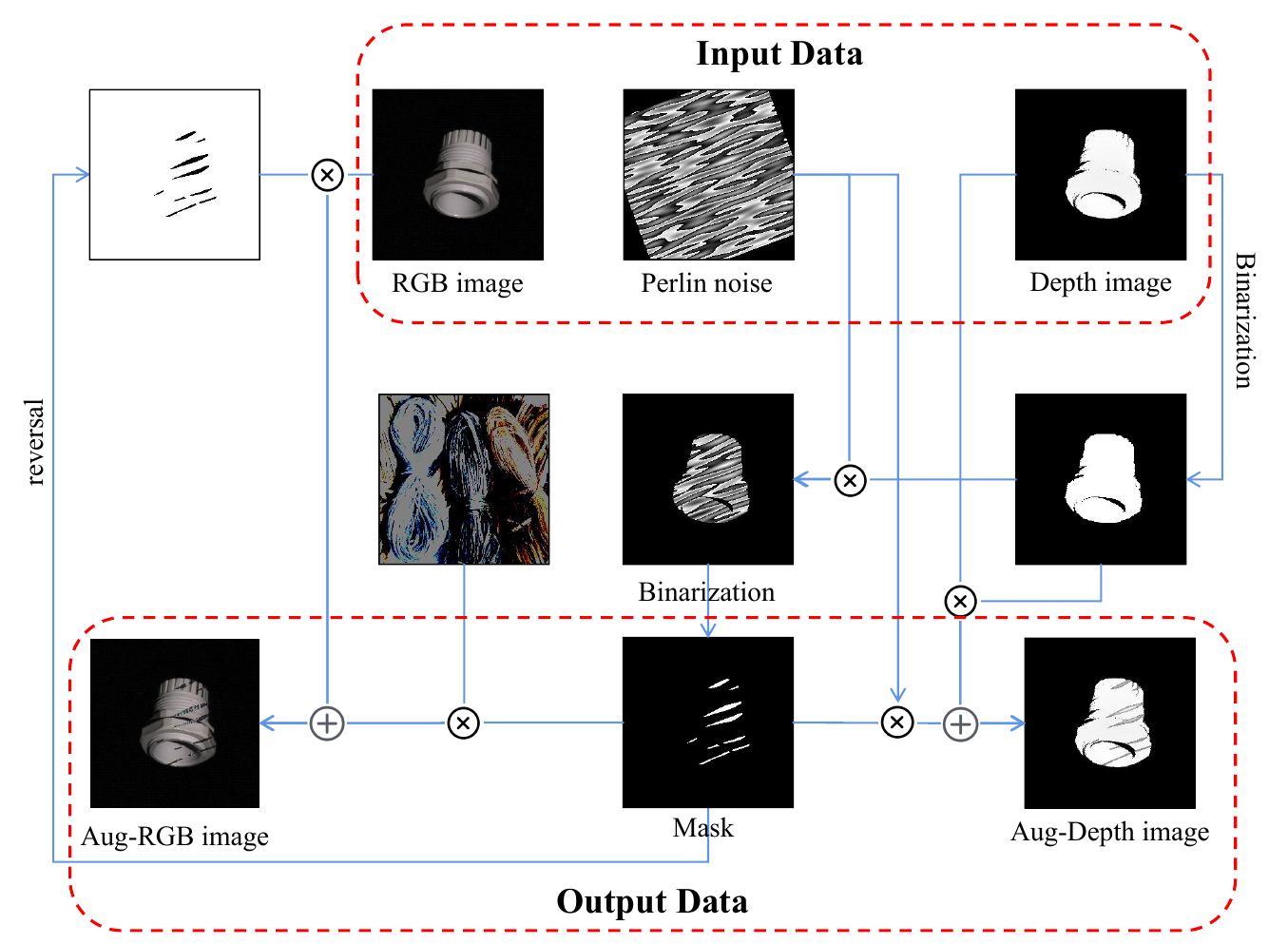}
	\caption{Illustration of anomaly generation processing.
	}\label{fig:anomaly_generation}
\end{figure}

The framework of the simulator is depicted in Figure~\ref{fig:anomaly_generation}. We generate a foreground mask on the original depth image and apply a mask operation on the randomly generated Berlin noise figure. Our empirical evaluation reveals that only adding foreground noise exclusively assists the network in recognizing the noise on the foreground object rapidly. Then, the Berlin noise map undergoes binarization to produce positive and negative mask maps. Both random and original RGB images undergo weighting, along with the Berlin noise map and depth image. Finally, the resulting outputs include RGB and depth images with anomalies and masks.

In the reconstruction task, we used the classic $L_{2}$ to reduce perceptual differences in RGB image reconstruction, we used the SSIM loss function in the RGB image reconstruction task. In our experiment, it is also found that spatial variation and multi-scale features have limited and even negative effects on depth images. Therefore, the final image reconstruction loss function should be:
\begin{equation}
\begin{aligned}
L_{rec}(I, I_r) & = L_{rec}^{RGB}(I, I_r) + L_{rec}^{depth}(I, I_r) \\
& = \lambda_1L_{SSIM}^{RGB}(I, I_r)+\lambda_2 l_2^{RGB}(I, I_r)+\lambda_3 l_2^{depth}(I, I_r),
\end{aligned}
\end{equation}
where $\lambda_1$, $\lambda_2$, $\lambda_3$ are loss balancing hyper-parameters, and all are set to 1 in our experimental setting.

\subsubsection{Multi-modality Segmentation Network (MSN)}
\label{subsubsec:multi_seg_net}

The MSN evaluates the normality of each time slot $(H, W)$. Similar to DRAEM~\cite{zavrtanik2021draem}, the training set samples are processed by the simulator and the discriminator performs mask identification by identifying the input of enhanced images and reconstructed images. The difference between DRAEM and EasyNet refers to the \textit{supplementary materials}. EasyNet extracts multi-layer features evaluated by discriminators through an MRN. MSN utilizes multi-layer reconstruction features and enhanced image features, which come from our assumption that some features that deviate from the normal distribution will be removed gradually with the deepening of the multi-modality reconstruction network. By comparing the difference of eigenvalues before and after removal, the locations of anomalies can be obtained.

When extracting reconstruction features and enhancing image features of multiple layers, we mainly adopt the first three layers of shallow networks of MRN and the last three layers of reconstructed features and carry out up-sampling operations to adapt for features of multiple layers. Moreover, we conduct ablation experiments. As shown in Section~\ref{subsubsec:ablation_study_number}, experiments show that when two-layer features are adopted, both the accuracy and computing cost of the network are optimized.
We use a two-layer multi-layer perceptron (MLP) to process multi-layer scale features extracted from RGB and depth images respectively. Finally, we use another two-layer MLP structure to combine the features of the two modes and perform positive and negative discriminations for each pixel in the image. As shown in Section~\ref{subsec:main_result}, the proposed straightforward strategy is successful in reaching its goal.

\begin{figure}[th]
    \centering
    \includegraphics[width=0.85\linewidth]{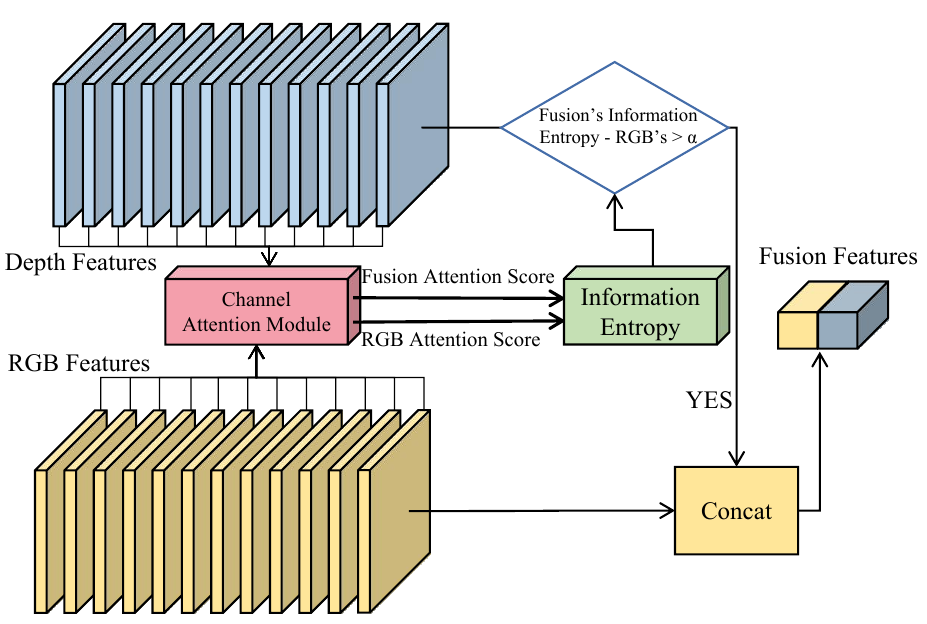}
	\caption{Illustration of Attention-based Information Entropy Fusion Module.
	}\label{fig:information_entropy_scheme}
\end{figure}

\subsubsection{Attention-based Information Entropy Fusion Module}

As noted in Section~\ref{subsec:challenges}, an anomaly may occur solely in pure RGB or depth images, or both. The direct combination of both features may diminish the overall performance of AD and lead to an inverse outcome. So we generate multi-channel self-attention scores from input features in the input layer of MSN's multi-layer perception module, as shown in Figure~\ref{fig:information_entropy_scheme}, We then compare the information entropy of the channel that integrates RGB and depth features with that of the channel integrating only pure RGB features. 
We hypothesize that the greater the information entropy of the channel attention score, the richer the feature knowledge it contains. If fusion features enhance the information gain beyond RGB features, it could positively affect the performance of the results. The experimental results presented in Section~\ref{subsubsec:Information_Entropy} provide support for our theory. The mathematical representation of this process is shown in Formula~\ref{con:f_fusion}.

\begin{equation}
\footnotesize
\begin{aligned}
F_{fusion} = 
\begin{cases}
F_{RGB}+F_{depth},&f_{IE}(F_{RGB}+F_{depth})>f_{IE}(F_{RGB})+\alpha\\
F_{RGB},&f_{IE}(F_{RGB}+f_{depth}) \leq f_{IE}(F_{RGB})+\alpha
\end{cases}
\end{aligned}
\label{con:f_fusion}
\end{equation}
\revised{where $F_{fusion}$ represents the features after fusion, $F_{RGB}$ represents the features of RGB, $F_{depth}$ represents the features of depth, $f_{IE}( \cdot)$ represents the function of calculating information entropy, and $\alpha$ represents the threshold adjustment factor.

When calculating the loss between the predicted mask and the ground truth mask, we use the Focal Loss~\cite{Lin2017FocalLF} function (Formula~\ref{con:focal_loss}), which could well solve the problem of sample imbalance in the single-class classification of pixels.}

\begin{equation}
\begin{aligned}
L_{focal}(M, M_{out}) = -\alpha_t(1-p_t)^\gamma log(p_t),
\end{aligned}
\label{con:focal_loss}
\end{equation}
where $\alpha_t$ is a scaling factor related to class $t$, $\gamma$ is an adjustable parameter, $p_t$ corresponds to the predicted classification of pixel points, the abnormal category is 1, and the normal category is 0.

To sum up, EasyNet optimization objectives and tasks are reconstruction loss and classification loss. Finally, the overall loss of the network during training is as follows:
\begin{equation}
\begin{aligned}
L_{all}(I, I_r) = & L_{rec}^{RGB}(I, I_r) + L_{rec}^{depth}(I, I_r) + L_{focal}(M, M_{out})\\
= & \lambda_1L_{SSIM}^{RGB}(I, I_r)+\lambda_2 l_2^{RGB}(I, I_r)\\
 & +\lambda_3 l_2^{depth}(I, I_r)+ \lambda_4 L_{focal}(M, M_{out}),
\end{aligned}
\end{equation}
where $\lambda_1$, $\lambda_2$, $\lambda_3$, and $\lambda_4$ are loss balancing hyper-parameters. \revised{Easynet aims to meet the objectives of optimizing anomaly detection and reconstruction tasks while training, so we optimize the above objectives by assigning weights to different losses. All four $\lambda$ are set to 1 in our experimental setting.}

\subsubsection{Algorithms}
\revised{The EasyNet is implemented as Algorithm~\ref{algorithm_easynet}. when training, images $I_{rgb}$ and $I_{depth}$ are enhanced by the anomaly generator $\Phi_{ag}$ to produce augmented images $A_{rgb}$ and $A_{depth}$ respectively. The Multi-modality Reconstruction Network $\Phi_{rec}$ extracts multi-scale features ($F_{rgb}, F_{depth} $) and generate reconstructed images ($R_{rgb}, R_{depth}$) from these augmented images and origin images. The Multi-modality Segmentation Network $\Phi_{seg}$ generates an anomaly score maps $M$ and $M_{rgb}$ by fusion and pure RGB features. When inferring, the function $\Phi_{ai}$ generates corresponding self-attention information entropy scores from both RGB and RGB-D channels to combine RGB and depth features.}

\begin{algorithm}
	\renewcommand{\algorithmicrequire}{\textbf{Input:}}
	\renewcommand{\algorithmicensure}{\textbf{Output:}}
	\caption{EasyNet pseudo-code}
	\label{algorithm_easynet}
	\begin{algorithmic}[1]
		\STATE \textbf{Input}: train dataloader $D_{train}$, test dataloader $D_{test}$, epochs
		\STATE \textbf{Output}: trained $\Phi_{rec}$ and $\Phi_{seg}$, $M$
        \STATE \textbf{Initialization ramdomly}:$\Phi_{rec}$ and $\Phi_{seg}$
        \STATE \textcolor{gray}{/*Training time*/}
        \FOR{$i = 0$ to epochs}
        \FOR{$I_{rgb}, I_{depth}, M_{gt} \leftarrow D_{train}$}
        \STATE $A_{rgb}, A_{depth} = \Phi_{ag}(I_{rgb}, I_{depth})$
        \STATE $F_{rgb}, F_{depth}, R_{rgb}, R_{depth} = \Phi_{rec}(A_{rgb}, A_{depth})$
        \STATE $F_{fusion} = Concat(F_{rgb}, F_{depth})$
        \STATE $M_{rgb} = \Phi_{seg}(F_{rgb})$
        \STATE $M = \Phi_{seg}(F_{fusion})$
        \STATE$L_{rgb} = \Phi_{loss}(R_{rgb}, I_{rgb}, M_{rgb}, M_{gt})$
        \STATE$L_{total} = \Phi_{loss}(R_{rgb}, R_{depth}, I_{rgb}, I_{depth}, M, M_{gt})$
        \STATE$L_{rgb}.backward, L_{total}.backward$
        \ENDFOR
        \ENDFOR
        \STATE \textcolor{gray}{/*Inference time*/}
        \FOR{$I_{rgb}, I_{depth}, M_{gt} \leftarrow D_{test}$}
        \STATE $A_{rgb}, A_{depth} = \Phi_{ag}(I_{rgb}, I_{depth})$
        \STATE $F_{rgb}, F_{depth}, R_{rgb}, R_{depth} = \Phi_{rec}(A_{rgb}, A_{depth})$
        \STATE $S_{rgb} = \Phi_{ai}(Feature_{rgb})$
        \STATE $S_{fusion} = \Phi_{ai}(F_{rgb}, F_{depth})$
        \IF{$S_{fusion}-S_{rgb} > \alpha$}
        \STATE $F_{fusion} = Concat(F_{rgb}, F_{depth})$
        \ELSE
        \STATE $F_{fusion} = F_{rgb}$
        \ENDIF
        \STATE $M = \Phi_{seg}(F_{fusion})$
        \ENDFOR
	\end{algorithmic}  
\end{algorithm}

\begin{table*}[t]
    \centering
    \caption{I-AUROC score for anomaly detection of MVTec 3D-AD. The best is in red and the second best is in blue.}
    \vspace{-10pt}
    \resizebox{0.88\textwidth}{!}{
    \begin{tabular}{l l | c c c c c c c c c c | c | c | c }
    \toprule[0.5mm]
        ~ & \textbf{Method} & \textbf{Bagel} & \textbf{Cable Gland} & \textbf{Carrot} & \textbf{Cookie} & \textbf{Dowel} & \textbf{Foam} & \textbf{Peach} & \textbf{Potato} & \textbf{Rope} & \textbf{Tire} & \textbf{Mean} & \thead{\textbf{Memory} \\\textbf{Bank Usage}} & \thead{\textbf{pretrained}\\\textbf{Model Usage}}\\ \hline
        \multirow{11}{*}{\textbf{\thead{Pure\\Depth}}} & Depth GAN~\cite{Bergmann2021TheM3}  & 0.530  & 0.376  & 0.607  & 0.603  & 0.497  & 0.484  & 0.595  & 0.489  & 0.536  & 0.521  & 0.523 & & \\ 
        ~ & Depth AE~\cite{Bergmann2021TheM3} & 0.468  & \blue{0.731}  & 0.497  & 0.673  & 0.534  & 0.417  & 0.485  & 0.549  & 0.564  & 0.546  & 0.546  & & \\ 
        ~ & Depth VM~\cite{Bergmann2021TheM3}  & 0.510  & 0.542  & 0.469  & 0.576  & 0.609  & 0.699  & 0.450  & 0.419  & 0.668  & 0.520  & 0.546 & &  \\ 
        ~ & Voxel GAN~\cite{Bergmann2021TheM3} & 0.383  & 0.623  & 0.474  & 0.639  & 0.564  & 0.409  & 0.617  & 0.427  & 0.663  & 0.577  & 0.537 & &  \\ 
        ~ & Voxel AE~\cite{Bergmann2021TheM3}  & 0.693  & 0.425  & 0.515  & 0.790  & 0.494  & 0.558  & 0.537  & 0.484  & 0.639  & 0.583  & 0.571 & &  \\ 
        ~ & Voxel VM~\cite{Bergmann2021TheM3}  & 0.750  & \red{0.747}  & 0.613  & 0.738  & 0.823  & 0.693  & 0.679  & 0.652  & 0.609  & \blue{0.690}  & 0.699  & & \\ 
        ~ & 3D-ST~\cite{Bergmann2022AnomalyDI} & 0.862  & 0.484  & 0.832  & 0.894  & 0.848  & 0.663  & 0.763  & 0.687  & \red{0.958}  & 0.486  & 0.748 &  &\Checkmark   \\ 
        ~ & PatchCore+FPFH~\cite{Horwitz2022AnEI} & 0.825  & 0.551  & \blue{0.952}  & 0.797  & \blue{0.883}  & 0.582  & 0.758  & 0.889  & 0.929  & 0.653  & 0.782 &\Checkmark &    \\ 
        ~ & AST~\cite{rudolph2022asymmetric} & \blue{0.881}  & 0.576  & \red{0.965}  & \blue{0.957}  & 0.679  & \red{0.797}  & \red{0.990}  & \blue{0.915}  & \blue{0.956}  & 0.611  & \blue{0.833} &  &\Checkmark \\ 
        ~ & M3DM~\cite{Wang2023MultimodalIA} & \red{0.941}  & 0.651  & \red{0.965}  & \red{0.969}  & \red{0.905}  & \blue{0.760}  & \blue{0.880}  & \red{0.974}  & 0.926  & \red{0.765}  & \red{0.874} &\Checkmark &\Checkmark \\ 
        ~ & \textbf{EasyNet(ours)} & 0.735  & 0.678  & 0.747  & 0.864  & 0.719  & 0.716  & 0.713  & 0.725  & 0.885  & 0.687  & 0.747 &   &   \\ 
        \hline
        \multirow{8}{*}{\textbf{\thead{Pure\\RGB}}} & DifferNet~\cite{Rudolph2020SameSB} & 0.859  & 0.703  & 0.643  & 0.435  & 0.797  & 0.790  & 0.787  & 0.643  & 0.715  & 0.590  & 0.696 &  &\Checkmark \\ 
        ~ & PADiM~\cite{Defard2020PaDiMAP}  & \blue{0.975}  & 0.775  & 0.698  & 0.582  & 0.959  & 0.663  & 0.858  & 0.535  & 0.832  & 0.760  & 0.764  &\Checkmark  &\Checkmark \\ 
        ~ & PatchCore~\cite{Roth2021TowardsTR}  & 0.876  & 0.880  & 0.791  & 0.682  & 0.912  & 0.701  & 0.695  & 0.618  & 0.841  & 0.702  & 0.770 &\Checkmark &\Checkmark \\ 
        ~ & STEPM~\cite{Wang2021StudentTeacherFP}  & 0.930  & 0.847  & 0.890  & 0.575  & 0.947  & 0.766  & 0.710  & 0.598  & 0.965  & 0.701  & 0.793 &  &\Checkmark \\ 
        ~ & CS-Flow~\cite{Gudovskiy2021CFLOWADRU}  & 0.941  & 0.930  & 0.827  & 0.795  & \blue{0.990}  & 0.886  & 0.731  & 0.471  & \blue{0.986}  & 0.745  & 0.830  &  &\Checkmark \\ 
        ~ & AST~\cite{rudolph2022asymmetric} & 0.947  & 0.928  & 0.851  & \blue{0.825}  & 0.981  & \red{0.951}  & \blue{0.895}  & 0.613  & \red{0.992}  & \blue{0.821}  & \blue{0.880} &   &\Checkmark \\ 
        ~ & M3DM~\cite{Wang2023MultimodalIA} & 0.944  & 0.918  & 0.896  & 0.749  & 0.959  & 0.767  & \red{0.919}  & 0.648  & 0.938  & 0.767  & 0.850 &\Checkmark &\Checkmark \\ 
        ~ & SPADE~\cite{Cohen2020SubImageAD} & 0.771 & 0.793 & 0.760 & 0.531 & 0.848 & 0.683 & 0.646 & 0.460 & 0.879 & 0.502 & 0.687 &\Checkmark &\Checkmark \\ 
        ~ & FastFlow~\cite{Yu12021FastFlowUA} & 0.624 & 0.472 & 0.654 & 0.694 & 0.501 & 0.667 & 0.595 & 0.632 & 0.816 & 0.731 & 0.639 &  &\Checkmark \\ 
        ~ & RD4AD~\cite{Deng2022AnomalyDV} & \blue{0.975} & \blue{0.987} & \red{0.943} & 0.575 & \red{0.999} & 0.830 & 0.863 & 0.618 & 0.984 & \red{0.899} & 0.867 &  &\Checkmark \\ 
        ~ & STPM~\cite{Wang2021StudentTeacherFP} & 0.899 & 0.706 & 0.796 & 0.486 & 0.512 & 0.678 & 0.502 & \blue{0.666} & 0.962 & 0.581 & 0.679 &  &\Checkmark \\ 
        ~ & \textbf{EasyNet(ours)} & \red{0.982}  & \red{0.992}  & \blue{0.917}  & \red{0.953}  & 0.919  & \blue{0.923 } & 0.840  & \red{0.785}  & \blue{0.986}  & 0.742  & \red{0.904} &   &   \\ 
        \hline
        \multirow{11}{*}{\textbf{\thead{RGB+\\Depth}}} & Depth GAN~\cite{Bergmann2021TheM3}  & 0.538  & 0.372  & 0.580  & 0.603  & 0.430  & 0.534  & 0.642  & 0.601  & 0.443  & 0.577  & 0.532 &   &   \\ 
        ~ & Depth AE~\cite{Bergmann2021TheM3} & 0.648  & 0.502  & 0.650  & 0.488  & 0.805  & 0.522  & 0.712  & 0.529  & 0.540  & 0.552  & 0.595 &   &  \\ 
        ~ & Depth VM~\cite{Bergmann2021TheM3}  & 0.513  & 0.551  & 0.477  & 0.581  & 0.617  & 0.716  & 0.450  & 0.421  & 0.598  & 0.623  & 0.555 &  &  \\ 
        ~ & Voxel GAN~\cite{Bergmann2021TheM3} & 0.680  & 0.324  & 0.565  & 0.399  & 0.497  & 0.482  & 0.566  & 0.579  & 0.601  & 0.482  & 0.517 &  &  \\ 
        ~ & Voxel AE~\cite{Bergmann2021TheM3}  & 0.510  & 0.540  & 0.384  & 0.693  & 0.446  & 0.632  & 0.550  & 0.494  & 0.721  & 0.413  & 0.538 &  &  \\ 
        ~ & Voxel VM~\cite{Bergmann2021TheM3}  & 0.553  & 0.772  & 0.484  & 0.701  & 0.751  & 0.578  & 0.480  & 0.466  & 0.689  & 0.611  & 0.609 &  &  \\ 
        ~ & 3D-ST~\cite{Bergmann2022AnomalyDI} & 0.950  & 0.483  & \red{0.986}  & 0.921  & 0.905  & 0.632  & 0.945  & \red{0.988}  & 0.976  & 0.542  & 0.833 & & \Checkmark \\ 
        ~ & PatchCore+FPFH~\cite{Horwitz2022AnEI} & 0.918  & 0.748  & 0.967  & 0.883  & 0.932  & 0.582  & 0.896  & 0.912  & 0.921  & \red{0.886}  & 0.865 &\Checkmark & \\ 
        ~ & AST~\cite{rudolph2022asymmetric} & 0.983  & 0.873  & \blue{0.976}  & \blue{0.971}  & 0.932  & 0.885  & \red{0.974}  & \blue{0.981}  & \red{1.000}  & 0.797  & \blue{0.937} &  &\Checkmark \\ 
        ~ & M3DM~\cite{Wang2023MultimodalIA} & \red{0.994}  & \blue{0.909}  & 0.972  & \red{0.976}  & \red{0.960}  & \blue{0.942}  & \blue{0.973}  & 0.899  & 0.972  & \blue{0.850}  & \red{0.945} &\Checkmark &\Checkmark \\ 
        ~ & \textbf{EasyNet(ours)} & \blue{0.991}  & \red{0.998}  & 0.918  & 0.968  & \blue{0.945}  & \red{0.945}  & 0.905  & 0.807  & \blue{0.994}  & 0.793  & 0.926 & &  \\ 
         \bottomrule[0.5mm]
    \end{tabular}
    }
    \label{tab:mvtec-3d-benchmark}
\end{table*}

\section{Experiments}
\subsection{Experimental Details}

\subsubsection{Datasets}

\revised{We mainly used MVTec 3D-AD~\cite{Bergmann2021TheM3} and Eyescandies~\cite{bonfiglioli2022eyecandies} data sets in the experiment. MVTec 3D-AD dataset is the data set of the real scene, and Eyescandies is the data set of the virtual synthesis. More detailed introduction to these two datasets please refer to the \textit{supplementary materials}.}

\subsubsection{Evaluation Metrics}

Due to the unsupervised experimental setting, the common evaluation metrics we used include Area Under the Receiver Operating Characteristic Curve (AUROC) (I-AUROC and P-AUROC) and the Area Under the Precision-Recall curve (AUPR/AP), the explanation of I-AUROC and P-AUROC please refer to the \textit{supplementary materials}.

\begin{table*}[th]
    \centering
    \caption{I-AUROC score for anomaly detection of all categories of Eyescandies. The best is in red and the second best is in blue.}
        \vspace{-10pt}
    \resizebox{0.88\textwidth}{!}{
    \begin{tabular}{l l | c c c c c c c c c c | c | c | c }
    \toprule[0.5mm]
        ~ & \textbf{Method} & \textbf{\thead{Candy \\Cane}} & \textbf{\thead{Chocolate\\Cookie}} & \textbf{\thead{Chocolate\\Cookie}} & \textbf{Confetto} & \textbf{\thead{Gummy\\Bear}} & \textbf{\thead{Hazelnut\\Truffle}} & \textbf{\thead{Licorice\\Sandwich}} &\textbf{ Lollipop} & \textbf{\thead{Marsh-\\mallow}} & \textbf{\thead{Peppermint\\Candy}} & \textbf{Mean} & \textbf{\thead{Memory \\Bank usage}} & \textbf{\thead{pretrained\\Model Usage}}\\ \hline
        \multirow{5}{*}{\textbf{\thead{Pure\\Depth}}} & Raw~\cite{Horwitz2022AnEI} & \blue{0.654} & 0.510 & 0.563 & 0.451 & 0.433 & 0.454 & 0.472 & 0.515 & 0.626 & 0.366 & 0.504 &\Checkmark & \\ 
         ~ &HoG~\cite{Horwitz2022AnEI} & 0.653 & 0.510 & 0.470 & 0.723 & \blue{0.728} & \blue{0.520} & 0.717 & 0.667 & 0.699 & 0.742 & 0.643 &\Checkmark &\\ 
         ~ &SIFT~\cite{Horwitz2022AnEI} & 0.589 & 0.582 & 0.683 & \red{0.885} & 0.663 & 0.480 & \blue{0.778} & 0.702 & \blue{0.746} & \red{0.790} & 0.690 &\Checkmark &\\ 
         ~ &FPFH~\cite{Horwitz2022AnEI} & \red{0.670} & \blue{0.710} & \red{0.805} & \blue{0.806} & \red{0.748} & 0.515 & \red{0.794} & \red{0.757} & \red{0.765} & \blue{0.757} & \red{0.733} &\Checkmark &\\ 
        ~ & \textbf{EasyNet(ours)} & 0.629  & \red{0.716}  & \blue{0.768}  & 0.731  & 0.660  & \red{0.710}  & 0.712  & \blue{0.711}  & 0.688 & 0.731 & \blue{0.706} & &\\ 
        \hline
        \multirow{7}{*}{\textbf{\thead{Pure\\RGB}}}
         ~ & GANomaly~\cite{Akay2018GANomalySA} & 0.485  & 0.512  & 0.532  & 0.504  & 0.558  & 0.486  & 0.467  & 0.511  & 0.481  & 0.528  & 0.507 & & \\ 
        ~ & DFKDE~\cite{anomalib} & 0.539  & 0.577  & 0.482  & 0.548  & 0.541  & 0.492  & 0.524  & 0.602  & 0.658  & 0.591  & 0.555 &  &\Checkmark \\ 
         ~ & DFM~\cite{Ahuja2019ProbabilisticMO} & 0.532  & 0.776  & 0.624  & 0.675  & 0.681  & 0.596  & 0.685  & 0.618  & 0.964  & 0.770  & 0.692 &  &\Checkmark \\ 
         ~ & STEPM~\cite{Wang2021StudentTeacherFP}  & \blue{0.551}  & 0.654  & 0.576  & 0.784  & \blue{0.737}  & \blue{0.790}  & 0.778  & 0.620  & 0.840  & 0.749  & 0.708  &  &\Checkmark \\ 
         ~ & PaDiM~\cite{Defard2020PaDiMAP} & 0.531  & 0.816  & \blue{0.821}  & \blue{0.856}  & \red{0.826}  & 0.727  & \blue{0.784}  & 0.665  & \red{0.987}  & \blue{0.924}  & \blue{0.794} & \Checkmark  &\Checkmark\\ 
        ~ & AutoEncoder~\cite{Bonfiglioli2022TheED} &0.527  & \blue{0.848}  & 0.772  & 0.734  & 0.590  & 0.508  & 0.693  & \blue{0.760}  & 0.851  & 0.730  & 0.701  & & \\ 
        ~ & \textbf{EasyNet(ours)} & \red{0.723}  & \red{0.925}  & \red{0.849}  & \red{0.966}  & 0.705  & \red{0.815}  & \red{0.806}  & \red{0.851}  & \blue{0.975}  & \red{0.960}  & \red{0.858} & & \\ \hline
        \multirow{3}{*}{\textbf{\thead{RGB+\\Depth}}} & AutoenEoder~\cite{Bonfiglioli2022TheED} &0.529  & 0.861  & 0.739  & 0.752  & 0.594  & 0.498  & 0.679  & 0.651  & 0.838  & 0.750  & 0.689 & &\\
        ~ & PatchCore+FPFH~\cite{Horwitz2022AnEI} & \blue{0.606}  & \blue{0.904}  & \blue{0.792}  & \blue{0.939}  & \red{0.720}  & \blue{0.563}  & \blue{0.867}  & \blue{0.860}  & \red{0.992}  & \blue{0.842}  & \blue{0.809} &\Checkmark &\Checkmark\\
        ~ & \textbf{EasyNet(ours)} & \red{0.737}  &  \red{0.934}  &  \red{0.866}  &  \red{0.966} &  \blue{0.717}  &  \red{0.822}  &  \red{0.847}  &  \red{0.863} & \blue{0.977}  &  \red{0.960}  & \red{0.869}  & &  \\  
         \bottomrule[0.5mm]
    \end{tabular}
    }
    \label{tab:eyescandies-benchmark}
\end{table*}

\begin{table*}[th]
    \centering
    \caption{Ablation studies on an attention-based information entropy fusion module. The best is in red and the second best is in blue. }
        \vspace{-10pt}
    \resizebox{0.85\textwidth}{!}{
    \begin{tabular}{l|l | c c c c c c c c c c c}
    \toprule[0.5mm]
    \multirow{4}{*}{\textbf{MVTec 3D-AD}} & \textbf{Evaluation} & \textbf{Bagel} & \textbf{Cable gland} & \textbf{Carrot} & \textbf{Cookie} & \textbf{Dowel} & \textbf{Foam} & \textbf{Peach} & \textbf{Potato} & \textbf{Rope} & \textbf{Tire} & \textbf{Mean} \\ 
    \cline{2-13}
         & Gate Close & \blue{0.982}  & 0.992  & \blue{0.917}  & \blue{0.953}  & 0.919  & 0.923  & 0.840  & \blue{0.785}  & \blue{0.986}  & \blue{0.742}  & 0.904  \\ 
        & Gate Open & 0.974  & \blue{0.996}  & 0.914  & \red{0.968}  & \blue{0.941}  & \blue{0.938}  & \blue{0.882}  & 0.781  & 0.982  & \red{0.793}  & \blue{0.917}  \\ 
        &Gate Control & \red{0.991}  & \red{0.998}  & \red{0.918}  & \red{0.968}  & \red{0.945}  & \red{0.945}  & \red{0.905}  & \red{0.807}  & \red{0.994}  & \red{0.793}  & \red{0.926}  \\ 
        \Xhline{1.5pt}
   \multirow{4}{*}{\textbf{Eyescandies}} & \textbf{Evaluation} & \textbf{\thead{Candy \\Cane}} & \textbf{\thead{Chocolate\\Cookie}} & \textbf{\thead{Chocolate\\Cookie}} & \textbf{Confetto} & \textbf{\thead{Gummy\\Bear}} & \textbf{\thead{Hazelnut\\Truffle}} & \textbf{\thead{Licorice\\Sandwich}} &\textbf{ Lollipop} & \textbf{\thead{Marsh-\\mallow}} & \textbf{\thead{Peppermint\\Candy}} & \textbf{Mean} \\ 
   \cline{2-13}
         & Gate Close & \blue{0.723}  & \blue{0.925}  & \blue{0.849}  & \red{0.966 } & \blue{0.705}  & \blue{0.815}  & 0.806  & \blue{0.851}  & \blue{0.975}  & \red{0.960}  & \blue{0.857}  \\  
        & Gate Open & 0.722  & 0.919  & 0.827  & \blue{0.945}  & 0.685  & 0.813  & \blue{0.846}  & 0.850  & \blue{0.975}  & \blue{0.959}  & 0.854  \\ 
        & Gate Control & \red{0.737}  & \red{0.934} & \red{0.866}  & \red{0.966}  & \red{0.717}  & \red{0.822}  & \red{0.847}  & \red{0.863}  & \red{0.977}  & \red{0.960}  & \red{0.869}\\ 
        \bottomrule[0.5mm]
    \end{tabular}
    }   
    \label{tab:gate-abalation-study}
\end{table*}

\begin{table*}[th]
    \centering
    \caption{The ablation study for the number of fusion layers. The best is in red and the second best is in blue. }
        \vspace{-10pt}
    \resizebox{0.85\textwidth}{!}{
    \begin{tabular}{{l|l c c c c c c c c c c c}}
    \toprule[0.5mm]
        \textbf{RGB+Depth} & \textbf{Evaluation} & \textbf{Bagel} & \textbf{Cable gland} & \textbf{Carrot} & \textbf{Cookie} & \textbf{Dowel} & \textbf{Foam} & \textbf{Peach} & \textbf{Potato} & \textbf{Rope} & \textbf{Tire} & \textbf{Mean} \\ \hline
        
        \multirow{3}{*}{Image AUC} & 1 layer & \blue{0.967} & 0.981 & \blue{0.912} & \blue{0.890} & 0.901 & 0.908 & \blue{0.763} & \blue{0.717} & \red{1.000} & 0.731 & 0.877 \\ 
        ~& 2 layers & \red{0.974} & \blue{0.996} & \red{0.914} & \red{0.968} & \blue{0.941} & \red{0.938} & \red{0.882} & \red{0.781} & \blue{0.982} & \blue{0.793} & \red{0.917} \\
        ~& 3 layers & 0.964 & \red{1.000} & 0.884 & 0.889 & \red{0.967} & \blue{0.932} & 0.734 & 0.697 & \red{1.000} & \red{0.866} & \blue{0.893} \\ \hline

        \multirow{3}{*}{Image AP} & 1 layer & \blue{0.992} & 0.995 & \red{0.982} & \blue{0.969} & 0.976 & 0.975 & \blue{0.915} & 0.873 & \red{1.000} & 0.906 & 0.958 \\ 
        ~& 2 layers & \red{0.994} & \blue{0.999} & \blue{0.981} & \red{0.991} & \blue{0.984} & \red{0.984} & \red{0.966} & \red{0.930} & \blue{0.992} & \blue{0.927} & \red{0.975} \\ 
        ~& 3 layers & 0.991 & \red{1.000} & 0.975 & 0.966 & \red{0.992} & \blue{0.982} & 0.903 & \blue{0.916} & \red{1.000} & \red{0.963} & \blue{0.969} \\ \hline

        \multirow{3}{*}{Pixel AUC} & 1 layer & 0.875 & \blue{0.861} & \blue{0.963} & \blue{0.678} & 0.716 & \red{0.998} & \blue{0.969} & \blue{0.921} & 0.928 & \blue{0.861} & \blue{0.877} \\ 
        ~& 2 layers & \red{0.935} & \red{0.941} & \red{0.971} & \red{0.897} & \red{0.885} & \blue{0.997} & \red{0.992} & 0.888 & \blue{0.955} & 0.728 & \red{0.919} \\ 
        ~& 3 layers & \blue{0.904} & 0.717 & 0.836 & 0.651 & \blue{0.809} & \blue{0.997} & 0.914 & \red{0.942} & \red{0.986} & \red{0.970} & 0.873 \\ \hline

        \multirow{3}{*}{Pixel AP} & 1 layer & 0.020 & 0.025 & \blue{0.097} & \blue{0.015} & 0.026 & \red{0.803} & \blue{0.081} & \blue{0.032} & 0.133 & \red{0.163} & \blue{0.139} \\ 
        ~& 2 layers & \blue{0.039} & \red{0.062} & \red{0.188} & \red{0.025} & \blue{0.034} & 0.562 & \red{0.298} & \red{0.034} & \blue{0.144} & 0.031 & \red{0.142} \\ 
        ~& 3 layers & \red{0.041} & \blue{0.043} & 0.084 & 0.008 & \red{0.121} & \blue{0.648} & 0.017 & 0.014 & \red{0.249} & \blue{0.134} & 0.136 \\ 
        \bottomrule[0.5mm]
    \end{tabular}
    }
    \label{table:ablation_number_layer}
\end{table*}

\subsection{Experimental Results and Analysis}
\label{subsec:main_result}

\subsubsection{RGB+Depth on MVTec 3D-AD and Eyescandies}

Table~\ref{tab:mvtec-3d-benchmark} and Table~\ref{tab:eyescandies-benchmark} clearly demonstrate that EasyNet achieves the state-of-the-art performance on MVTec 3D-AD and Eyescandies without using a pretrained model and memory bank. Specifically, EasyNet outperforms AutoEncoder~\cite{Bonfiglioli2022TheED} and PatchCore+FPFH~\cite{Horwitz2022AnEI} in RGB+Depth setting of Eyescandies by a large margin, 20.7\% and 6.9\%, respectively. For MVTec 3D-AD, AST~\cite{rudolph2022asymmetric} and M3DM~\cite{Wang2023MultimodalIA} are the cutting-edge models in MVTec 3D-AD. However, both of them use large pretrained models or memory banks. In specific, M3DM uses two pretrained models (Point Transformer and Vision Transformer) to extract the features from depth images and RGB images. In addition, M3DM employs two large memory banks (average 6.098 GB) to store the features from depth images and RGB images. Due to strict storage limitations in practice, M3DM cannot be perfectly fit in real-world applications. Although PatchCore+FPFH~\cite{Horwitz2022AnEI} has a relatively small memory footprint (249.260 MB on average), the actual performance is not as good as Easynet. The memory bank size of M3DM and PatchCore+FPFH on the MVTec 3D-AD can be found in \textit{supplementary materials}. AST also adopts two EfficientNet-B5 as the feature extractors for depth images and RGB images, which violate the storage limitation in IM. Moreover, according to Table~\ref{tab:inference-accuracy}, massive usage of pretrained models will slow down the inference speed, which cannot meet the requirement of IM. Furthermore, the performance gap among EasyNet, AST and M3DM is very small, 2.1\% and 1.2\%. Hence, EasyNet is the best 3D-AD model to meet all the demands of IM.

\subsubsection{Pure RGB Performance}

In real-world applications, the limitation of the depth sensor is very large since the effective distance range of the depth sensor is 3 meters. In addition, most depth sensors are easily affected by the lighting condition. Hence, as for simulating the failure of the depth sensor, we conducted the experiment using only RGB images as the input. In the pure RGB branch of Table~\ref{tab:mvtec-3d-benchmark} and Table~\ref{tab:eyescandies-benchmark}, EasyNet achieves state-of-the-art performance in I-AUROC in the pure RGB track. In particular, EasyNet outperforms in pure RGB of Eyescandies by a large margin, 15.7\% to AutoEncoder~\cite{Bonfiglioli2022TheED} and 6.4\% to PaDiM~\cite{Defard2020PaDiMAP}. Moreover, EasyNet gets 5.97\% better I-AUROC score than M3DM and 2.65\% better I-AUROC score than AST. In total, The performance of EasyNet is robust even though the depth sensor is a failure. 

\begin{figure*}[th]
    \centering
    \includegraphics[width=0.8\linewidth]{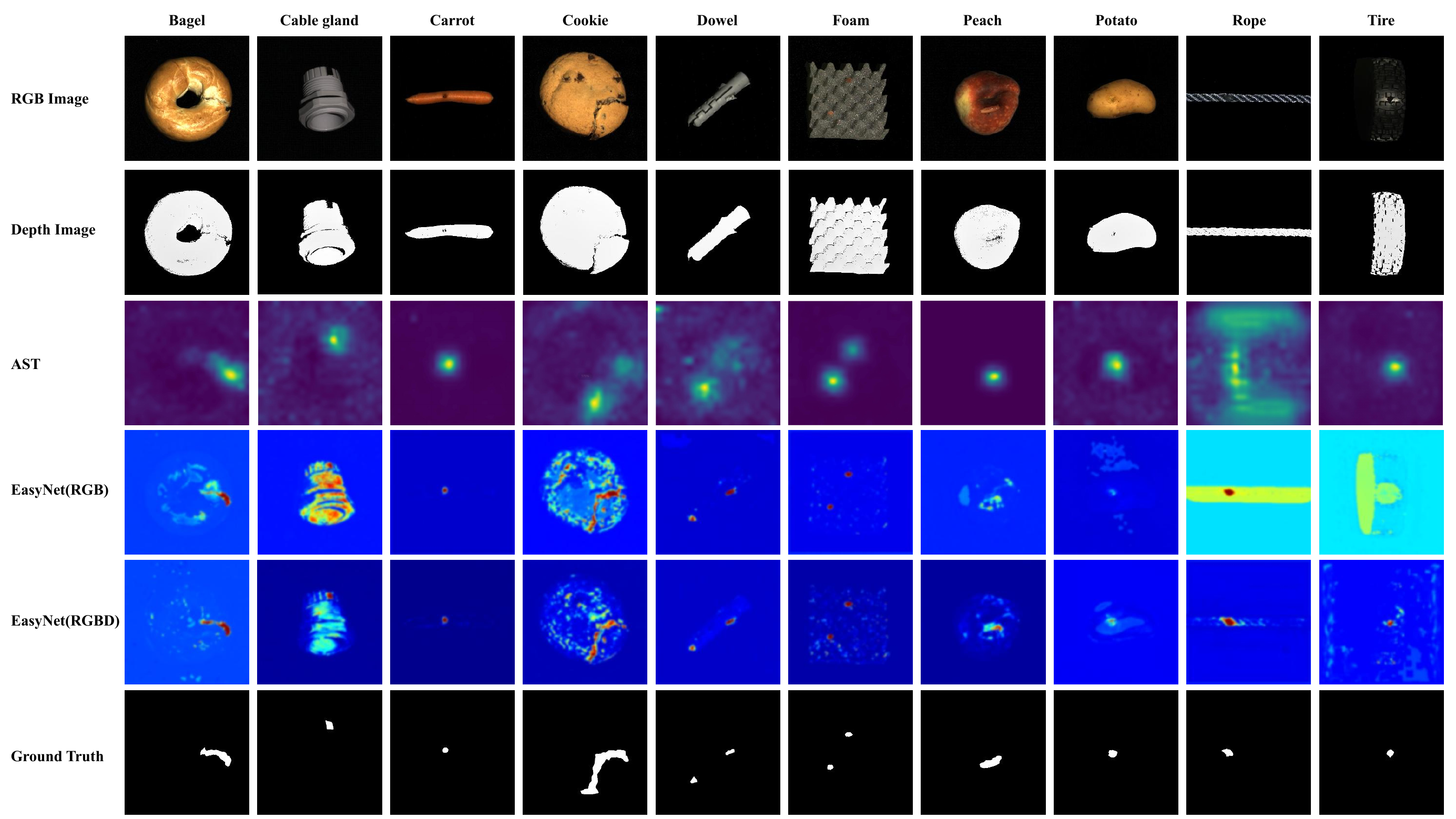}
	\caption{Visualizations on MVTec 3D-AD, which are obtained by AST~\cite{rudolph2022asymmetric}, EasyNet (RGB) and EasyNet (RGB-D), respectively. 
	}\label{fig:result_of_mvtec_3d_ad}
\end{figure*}

\subsubsection{Attention-based Information Entropy Fusion Module}
\label{subsubsec:Information_Entropy}
Table~\ref{tab:gate-abalation-study} clearly illustrates the effectiveness of our proposed attention-based information entropy fusion module in EasyNet. \revised{The gate network is the key to control multi-feature fusion.} We conduct the ablation studies on three options, Gate Close, Gate Open and Gate Control, respectively. Gate Close means that EasyNet only utilizes RGB images as the input and ignores depth information. Gate Open denotes that EasyNet uses both the RGB images and the depth images and combines their features for evaluation. Gate Control means that EasyNet adopts an attention-based information entropy fusion module to select depth features during the inference case. In Gate Open, we discover that depth information may work as the noise and degrade the total performance if we select both RGB features and depth features during inference, so we design an attention-based information entropy fusion module to select the feature for fusion, which can enhance the performance of all classes in MVTec 3D-AD and Eyescandies. 

\subsubsection{Ablation study on the number of fusion layers}
\label{subsubsec:ablation_study_number}

As described in Section~\ref{subsubsec:multi_seg_net}, the MSN is utilized for segmenting anomalies by fusing enhanced multilayer image features and reconstructed ones. Table~\ref{table:ablation_number_layer} presents that EasyNet performs optimally when incorporating features from only two layers. Specifically, the performance metrics of I-AUROC and P-AUROC are respectively improved by 4.36\% and 4.57\% with two layers compared to one layer, and they are further increased by 2.62\% and 5.27\% with three layers compared to one layer. These results indicate that the use of two feature layers can achieve the best performance in both anomaly detection and localization. In contrast, using three feature layers leads to performance degradation, which verifies our hypothesis. The deepening of the multi-modality reconstruction network can gradually eliminate some abnormal feature parts, whereas using three feature layers introduces more feature removal of abnormal parts, thereby leading to poor discriminator performance.

\subsubsection{Accuracy VS Inference Speed} \label{sec:accuracy_vs_inference}

\begin{wraptable}{r}{0.23\textwidth}
    \centering
    \caption{Inference abilities.}
        \vspace{-10pt}
    \scalebox{0.8}{
    \begin{tabular}{l| c c}
    \toprule[0.3mm]
        \textbf{Method} & \textbf{I-AUROC} & \textbf{FPS} \\ \hline
        BTF~\cite{horwitz2022back} & 0.865 & 27.92 \\ 
        AST~\cite{rudolph2022asymmetric} & 0.937 & 41.94 \\ 
       M3DM~\cite{Wang2023MultimodalIA} & 0.945 & 0.10 \\ 
        \textbf{EasyNet(ours)} & \textbf{0.926} & \textbf{94.55} \\ 
        \bottomrule[0.3mm]
    \end{tabular}}
    \label{tab:inference-accuracy}
\end{wraptable}
As we previously described in Section~\ref{sec:introduction}, inference speed is one of the important factors to be considered in IM. Since the real production lines need to check each product in real time. Table~\ref{tab:inference-accuracy} shows EasyNet obtains the fastest speed among the cutting-edge anomaly detection methods without sacrificing performance. In particular, EasyNet gets 125\% FPS better than AST and 93900\% FPS better than M3DM. Regarding performance, the performance gap among EasyNet, AST and M3DM is very small, 2.1\% and 1.2\%. Therefore, EasyNet is the most deployment-friendly 3D-AD method for IM.

\subsection{Visualization}

\revised{Figure~\ref{fig:result_of_mvtec_3d_ad} visualizes the performance of EasyNet on MVTec 3D-AD, demonstrating the effectiveness of the proposed method.
In Figure~\ref{fig:result_of_mvtec_3d_ad}, compared to existing 3D anomaly detection methods (AST~\cite{rudolph2022asymmetric}), EasyNet can significantly reduce false positive rates and achieve higher segmentation accuracy. Furthermore, the fusion method reduces false positives for anomalies such as \textit{cable gland}, compared to using only RGB images. Anomalies in \textit{peach} and \textit{potato} are also more clearly visible on depth images, indicating the importance of using depth image information in industrial AD. In addition, EasyNet is not affected by the domain gap between natural and industrial images and has a higher inference speed than existing methods. Note that we put the visualization results of EasyNet on Eyescandies in \textit{supplementary materials}.}

\section{Conclusions}
This paper addresses a promising and challenging task, i.e., deployment-friendly 3D-AD and proposes an easy but effective neural network (termed as EasyNet) to achieve competitive performance without using large pretrained models and memory banks. Specifically, as for getting rid of large pretrained models and memory banks, EasyNet employs MRN to implicitly detect and reconstruct the anomalies with semantically plausible anomaly-free content, while keeping the non-anomalous regions of the input image unchanged. Meanwhile, EasyNet proposes an MSN to produce an accurate anomaly segmentation map from the concatenated reconstructed RGB images and depth images and their original appearances. In the test phase, EasyNet adopts a self-attention information entropy score in the early fusion stage to select the informative depth features before fusing with RGB features. To this end, EasyNet achieves the fastest inference speed without sacrificing performance.

\section{Acknowledgments}
This work was supported by the National Key R\&D Program of China (Grant NO. 2022YFF1202903), the National Natural Science Foundation of China (Grant NO. 62122035, 62206122), and the Key-Area Research and Development Program of Guangdong Province (2020B0101130003).

\bibliographystyle{ACM-Reference-Format}
\balance
\bibliography{sample-base}

\newpage
\input{supplementary}

\end{document}

%% file: supplementary.tex






\section{supplementary}

\subsection{Difference between DRAEM and EasyNet}\label{sec:difference-DRAEM-EasyNet}
As stated in EasyNet, we learn from DRAEM's reconstruction and abnormal image generation methods, but the architecture between EasyNet and DRAEM is quite different. Firstly, the reconstruction network of EasyNet employs the multi-layer and multi-scale feature information before and after reconstruction, which is also highlighted in the ablation studies, while the counterpart of DREAM can only be segmented by the images before and after reconstruction. Secondly, our experiments find that 2-layer MLP usage is sufficient to effectively segment the abnormal region without the need to use a large U-Net like DREAM. Finally, EasyNet pays more attention to the fusion and segmentation of multi-modal features, while DREAM only focuses on RGB features.

\subsection{Datasets}
\textbf{MVTec 3D-AD~\cite{Bergmann2021TheM3}} includes ten categories and a total of 2,656 training samples along with 1,137 testing samples. The 3D scans in this dataset were acquired via a structured-light-powered industrial scanner that captured the x, y, and z coordinates of the target object. Additionally, RGB data is also collected at the same time for each point in the cloud. To process the 3D data accurately, it is crucial to remove all the background noise. A RANSAC algorithm is employed to estimate the background plane, ensuring that points within 0.005 distances were eliminated without disturbing the RGB data. However, their corresponding pixels in the RGB image were set to zero. This step minimized disturbances while enhancing the accuracy of anomaly detection. 

\textbf{Eyescandies~\cite{bonfiglioli2022eyecandies}} is a novel synthetic dataset comprising ten different categories of candies rendered in a controlled environment. Bonfiglioli \textit{et al.}~\cite{Bonfiglioli2022TheED} generated item instances through modeling software and collected relevant data. The dataset provides depth and RGB images in an industrial conveyor scenario. The ten categories of candies show different challenges, such as complex textures, self-occlusions, and specularities. By controlling the lighting conditions and parameters of a procedural rendering pipeline in the modeling software, the authors of the dataset produced datasets containing complex instances with varying conditions. Similar to MVTec 3D-AD, the training dataset only consists of normal samples, while the testing dataset consists of normal and abnormal samples.

\subsection{I-AUROC and P-AUROC}
I-AUROC notes Image-level AUROC and P-AUROC notes Pixel-level AUROC. Image-level AUROC is based on the area under the ROC curve of the entire image, where the horizontal axis represents a false positive rate and the vertical axis represents a true positive rate. Each point represents the performance of a model under different classification thresholds. Different points can be obtained by changing the image classification threshold to plot the entire ROC curve. Image-level AUROC is used to evaluate the classification quality of the overall image. Pixel-level AUROC reflects the segmentation accuracy of a model at the pixel level based on the area under the ROC curve of each pixel.

\subsection{Implementation Details}

This section presents the implementation details of our experiments. 

MRN uses the "UNet-like" structure as the primary network with intermediate skip operations subtracted primarily from the original UNet. The input image is resized to $256\times256$, and the abnormal and normal images are allocated according to a 1:1 ratio. The abnormal images are applied with Berlin noise~\cite{Zavrtanik2021DRMA} added on top of normal images.

For MSN, the two-layer MLP network is used to fuse different scale features of RGB and depth features. In the experiment of Section 4.2.3, the two layers MLPs network are employed to fuse different modal features. The input and output features of all the MLPs have the same size of $256 \times 256$. And a SE block~\cite{Hu2017SqueezeandExcitationN} is utilized for Attention-based Information Entroy Fusion Module to score channel attention for both modes.

The training process adopted the Adam optimizer with a learning rate of 0.002, which is dynamically adjusted twice, at $0.8 \times $ epochs and $ 0.9 \times $ epochs, with a multiplier factor of 0.2. The batch size is set to 8. Finally, we report the best anomaly detection results obtained after 800 training steps of MRN.

\subsection{Memory of Mainstream Methods}
Memory-based methods are also deployed in real-world applications, so the use of memory banks should not be restricted. Therefore, we calculate in Table 1 the memory size required by the current mainstream methods to use the RGBD method in the MVTec 3D-AD dataset.

\begin{table*}[!ht]
    \centering
    \scalebox{0.85}{
    \begin{tabular}{ c|c c c c c c c c c c c}
    \hline
        \textbf{Method} & \textbf{Bagel} & \textbf{Cable Gland} & \textbf{Carrot} & \textbf{Cookie} & \textbf{Dowel} & \textbf{Foam} & \textbf{Peach} & \textbf{Potato} & \textbf{Rope} & \textbf{Tire} & \textbf{Mean} \\ \hline
        PatchCore+FPFH~\cite{Horwitz2022AnEI}(MB) & 228.984 & 209.281 & 268.403 & 197.083 & 270.282 & 221.479 & 338.790 & 281.547 & 279.667 & 197.083 & 249.260 \\ 
        M3DM~\cite{Wang2023MultimodalIA}(GB) & 5.580 & 5.122 & 6.569 & 4.824 & 6.615 & 5.421 & 8.292 & 6.891 & 6.845 & 4.824 & 6.098 \\ \hline
        
    \end{tabular}
    }
    \caption{The size of the memory of mainstream methods using memory bank on the MVTec 3D-AD.}
    \label{tab:memory_bank_size}
\end{table*}

The M3DM method occupies a large amount of memory, resulting in a large amount of data transmission and loading time in the actual reasoning process. Although PatchCore+FPFH occupies a smaller memory, its performance is worse than EasyNet. The size of the memory bank is mainly affected by the training set, which indicates that they are using large memory sizes and hindering their practical application.

\subsection{AUPRO Score of MVTec 3D-AD}
In addition to I-AUROC, we also calculated the EasyNet model's AUPRO performance on the MvTec 3D-AD dataset, and Table~\ref{table:aupro_easynet} clearly shows that EasyNet achieves state-of-the-art performance on MvTec 3D-AD without the use of pre-trained models and memory banks and is slightly worse performance than 3D-ST using pre-trained models. In the experimental settings of Pure RGB and Pure Depth, our model does not take priority, but it also proves to a certain extent that our Attention-based Information Entropy Fusion Module plays a role, which blends the information of the two modes well.

\begin{table*}[!ht]
    \centering
    \scalebox{0.8}{
    \begin{tabular}{c c | c c c c c c c c c c | c | c | c }
    \toprule[0.8mm]
        ~ & \textbf{Method} & \textbf{Bagel} & \textbf{Cable Gland} & \textbf{Carrot} & \textbf{Cookie} & \textbf{Dowel} & \textbf{Foam} & \textbf{Peach} & \textbf{Potato} & \textbf{Rope} & \textbf{Tire} & \textbf{Mean} & \textbf{\thead{memeory \\bank use}} & \textbf{\thead{pretrain\\model use}}\\ \hline
        \multirow{9}{*}{\thead{Pure\\Depth}} & Depth GAN~\cite{Bergmann2021TheM3}  & 0.111  & 0.072  & 0.212  & 0.174  & 0.160  & 0.128  & 0.003  & 0.042  & 0.446  & 0.075  & 0.143 & &\\ 
        ~ & Depth AE~\cite{Bergmann2021TheM3} & 0.147  & 0.069  & 0.293  & 0.217  & 0.207  & 0.181  & 0.164  & 0.066  & 0.545  & 0.142  & 0.203 & &\\ 
        ~ & Depth VM~\cite{Bergmann2021TheM3}  & 0.280  & 0.374  & 0.243  & 0.526  & 0.485  & 0.314  & 0.199  & 0.388  & 0.543  & 0.385  & 0.374 & &\\ 
        ~ & Voxel GAN~\cite{Bergmann2021TheM3} & 0.440  & 0.453  & 0.875  & 0.755  & 0.782  & 0.378  & 0.392  & 0.639  & 0.775  & 0.389  & 0.583  & &\\ 
        ~ & Voxel AE~\cite{Bergmann2021TheM3}  &  0.260  & 0.341  & 0.581  & 0.351  & 0.502  & 0.234  & 0.351  & 0.658  & 0.015  & 0.185  & 0.348  & & \\ 
        ~ & Voxel VM~\cite{Bergmann2021TheM3}  & 0.453  & 0.343  & 0.521  & 0.697  & 0.680  & 0.284  & 0.349  & 0.634  & 0.616  & 0.346  & 0.492 & &\\ 
        ~ & FPFH~\cite{Horwitz2022AnEI} & \red{0.973}  & \red{0.879}  & \red{0.982}  & \red{0.906}  & \red{0.892}  & \blue{0.735}  & \red{0.977}  & \red{0.982}  & \red{0.956}  & \red{0.961}  & \red{0.924} &\Checkmark &    \\ 
        ~ & M3DM~\cite{Wang2023MultimodalIA} &  \blue{0.943}  & \blue{0.818}  & \blue{0.977}  & \blue{0.882}  & \blue{0.881}  & \red{0.743}  & \blue{0.958}  & \blue{0.974}  & \blue{0.950}  & \blue{0.929}  & \blue{0.906} &\Checkmark &\Checkmark \\ 
        ~ & \textbf{EasyNet(ours)} & 0.160  & 0.030  & 0.680  & 0.759  & 0.758  & 0.069  & 0.225  & 0.734  & 0.797  & 0.509  & 0.472 &   &   \\ 
        \hline
        \multirow{3}{*}{\thead{Pure\\RGB}} & PatchCore~\cite{Roth2021TowardsTR}  & \blue{0.901}  & \red{0.949}  & \red{0.928}  & \red{0.877}  & \blue{0.892}  & 0.563  & 0.904  & \red{0.932}  & \blue{0.908}  & \red{0.906}  & \red{0.876}  &\Checkmark &\Checkmark\\ 
        ~ & M3DM~\cite{Wang2023MultimodalIA} & \red{0.944}  & \blue{0.918}  & 0.896  & \blue{0.749}  & \red{0.959}  & \red{0.767}  & \red{0.919}  & 0.648  & \red{0.938}  & \blue{0.767}  & \blue{0.850} &\Checkmark &\Checkmark \\ 
        ~ & \textbf{EasyNet(ours)} & 0.751  & 0.825  & \blue{0.916}  & 0.599  & 0.698  & \blue{0.699}  & \blue{0.917}  & \blue{0.827}  & 0.887  & 0.636  & 0.776  &   &   \\ 
        \hline
        \multirow{8}{*}{\thead{RGB+\\Depth}} & Depth GAN~\cite{Bergmann2021TheM3}  & 0.421  & 0.422  & 0.778  & 0.696  & 0.494  & 0.252  & 0.285  & 0.362  & 0.402  & \red{0.631}  & 0.474 &   &   \\ 
        ~ & Depth AE~\cite{Bergmann2021TheM3} & 0.432  & 0.158  & 0.808  & 0.491  & \blue{0.841}  & 0.406  & 0.262  & 0.216  & 0.716  & 0.478  & 0.481 &   &  \\ 
        ~ & Depth VM~\cite{Bergmann2021TheM3}   & 0.388  & 0.321  & 0.194  & 0.570  & 0.408  & 0.282  & 0.244  & 0.349  & 0.268  & 0.331  & 0.335 &  &  \\ 
        ~ & Voxel GAN~\cite{Bergmann2021TheM3} & 0.664  & 0.620  & 0.766  & \blue{0.740}  & 0.783  & 0.332  & 0.582  & 0.790  & 0.633  & 0.483  & 0.639 &  &  \\ 
        ~ & Voxel AE~\cite{Bergmann2021TheM3}  & 0.467  & \blue{0.750}  & 0.808  & 0.550  & 0.765  & 0.473  & 0.721  & \blue{0.918}  & 0.019  & 0.170  & 0.564 &  &  \\ 
        ~ & Voxel VM~\cite{Bergmann2021TheM3}  & 0.510  & 0.331  & 0.413  & 0.715  & 0.680  & 0.279  & 0.300  & 0.507  & 0.611  & 0.366  & 0.471 &  &  \\ 
        ~ & 3D-ST~\cite{Bergmann2022AnomalyDI} & \red{0.950}  & 0.483  & \red{0.986}  & \red{0.921}  & \red{0.905}  & \blue{0.632}  & \red{0.945}  & \red{0.988}  & \red{0.976}  & \blue{0.542}  & \red{0.833} & & \Checkmark \\ 
        ~ & \textbf{EasyNet(ours)} & \blue{0.839}  & \red{0.864}  & \blue{0.951}  & 0.618  & 0.828  & \red{0.836}  & \blue{0.942}  & 0.889  & \blue{0.911}  & 0.528  & \blue{0.821} & &  \\ 
         \bottomrule[0.8mm]
    \end{tabular}
    }
    \caption{AUPRO score for anomaly detection of all categories of MVTec 3D-AD.}
    \label{tab:aupro_easynet}
\end{table*}

\subsection{Visualizations on Eyescandies}
\begin{figure*}[htbp]
    \centering
    \includegraphics[width=0.93\linewidth]{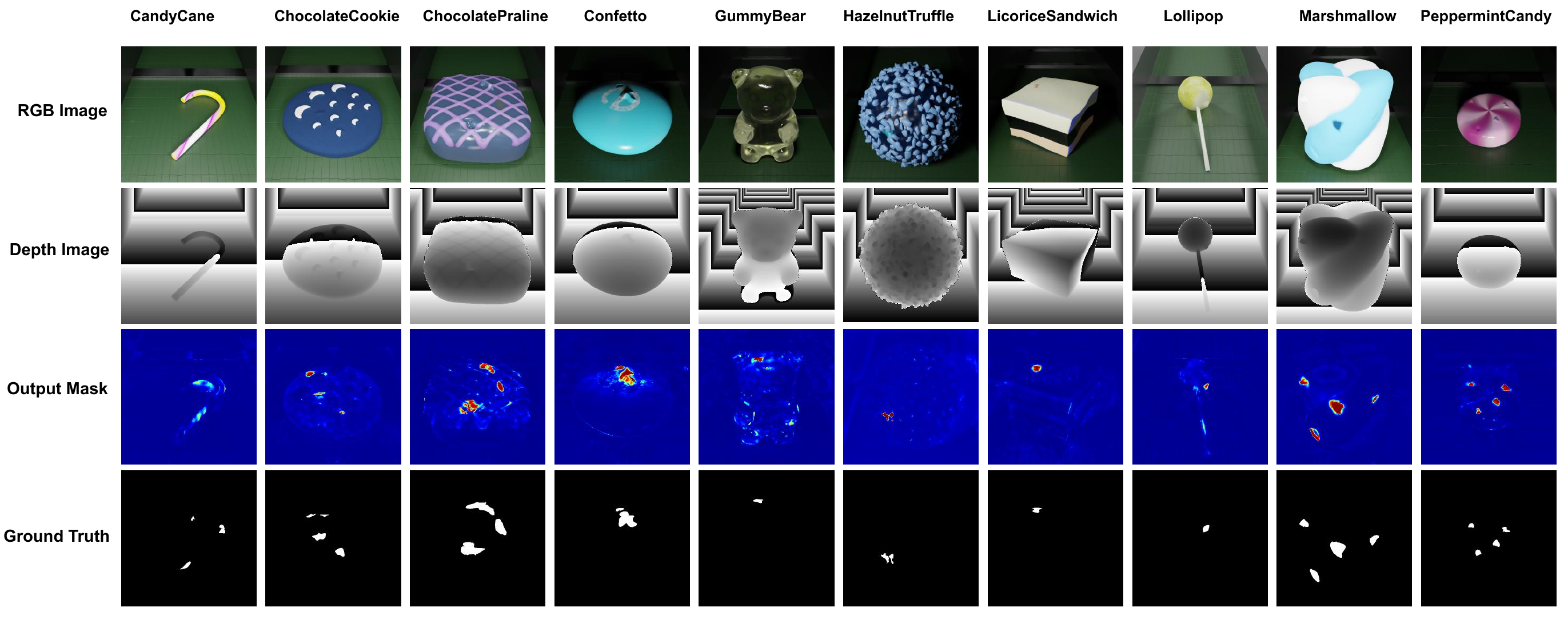}
	\caption{Visualizations on Eyescandies, which are obtained by EasyNet (RGB) and EasyNet (RGB-D). 
	}\label{fig:result_of_eyescandies}
\end{figure*}

Figure~\ref{fig:result_of_eyescandies} visualizes the performance of EasyNet on Eyescandies dataset, proving the effectiveness of the proposed method. In Figure~\ref{fig:result_of_eyescandies}, in the EasyNet dataset, although RGB images are interfered with by various lighting conditions, the anomaly heat map result of RGB image generated by EasyNet has a high coincidence degree with the ground truth. Compared with only RGB images, the contour of the anomaly heat map generated by EasyNet fusion method is clearer.

